\documentclass[english]{llncs}
\usepackage[T1]{fontenc}
\usepackage[latin9]{inputenc}
\usepackage{array}
\usepackage{float}
\usepackage{textcomp}
\usepackage{amstext}
\usepackage{graphicx}

\makeatletter

\newcommand{\lyxmathsym}[1]{\ifmmode\begingroup\def\b@ld{bold}
  \text{\ifx\math@version\b@ld\bfseries\fi#1}\endgroup\else#1\fi}

\providecommand{\tabularnewline}{\\}
\floatstyle{ruled}
\newfloat{algorithm}{tbp}{loa}
\providecommand{\algorithmname}{Algorithm}
\floatname{algorithm}{\protect\algorithmname}

\@ifundefined{showcaptionsetup}{}{%
 \PassOptionsToPackage{caption=false}{subfig}}
\usepackage{subfig}
\makeatother

\usepackage{babel}
\begin{document}

\title{ Machine learning on images using a string-distance}

\titlerunning{Image classification and clustering using a universal distance measure}

\author{Uzi Chester \and Joel Ratsaby\footnote{Corresponding author.}}

\authorrunning{U. Chester and J. Ratsaby} 
%
%
\institute{Electrical and Electronics Engineering Department, Ariel University of Samaria, ARIEL 40700,\\
\email{uzichester@gmail.com,  ratsaby@ariel.ac.il},\\ WWW home page:
\texttt{http://www.ariel.ac.il/sites/ratsaby/}
}


\tocauthor{Joel Ratsaby }
\maketitle
\begin{abstract}
We present a new method for image feature-extraction which is based
on representing an image by a finite-dimensional vector of distances
that measure how different the image is from a set of image prototypes.
We use the recently introduced Universal Image Distance (UID) \cite{RatsabyChesterIEEE2012}
to compare the similarity between an image and a prototype image.
The advantage in using the UID is the fact that no domain knowledge
nor any image analysis need to be done. Each image is represented
by a finite dimensional feature vector whose components are the UID
values between the image and a finite set of image prototypes from
each of the feature categories. The method is automatic since once
the user selects the prototype images, the feature vectors are automatically
calculated without the need to do any image analysis. The prototype
images can be of different size, in particular, different than the
image size. Based on a collection of such cases any supervised or
unsupervised learning algorithm can be used to train and produce an
image classifier or image cluster analysis. In this paper we present
the image feature-extraction method and use it on several supervised
and unsupervised learning experiments for satellite image data.
\end{abstract}

\section{Introduction}

Image classification research aims at finding representations of images
that can be automatically used to categorize images into a finite
set of classes. Typically, algorithms that classify images require
some form of pre-processing of an image prior to classification. This
process may involve extracting relevant features and segmenting images
into sub-components based on some prior knowledge about their context
\cite{canty2007image,lilesand2008remote}.

In \cite{RatsabyChesterIEEE2012} we introduced a new distance function,
called Universal Image Distance (UID), for measuring the distance
between two images. The UID first transforms each of the two images
into a string of characters from a finite alphabet and then uses the
string distance of \cite{Sayood2003} to give the distance value between
the images. According to \cite{Sayood2003} the distance between two
strings $x$ and $y$ is a normalized difference between the complexity
of the concatenation $xy$ of the strings and the minimal complexity
of each of $x$ and $y$. By complexity of a string $x$ we mean the
Lempel-Ziv complexity \cite{LZ76}.

In the current paper we use the UID to create a finite-dimensional
representation of an image. The $i^{th}$ component of this vector
is like a feature that measures how different the image is from the
$i^{th}$ image prototype. One of the advantages of the UID is that
it can compare the distance between two images of different sizes
and thus the prototypes which are representative of the different
feature categories may be relatively small. For instance, the prototypes
of an \emph{urban} category can be small images of size $45\times17$
pixels of various parts of cities.

In this paper we introduce a process to convert the image into a labeled
case (feature vector). Doing this systematically for a set of images
each labeled by its class yields a data set which can be used for
training any supervised and unsupervised learning algorithms. After
describing our method in details we report on the accuracy results
of several classification-learning algorithms on such data. As an
example, we apply out method to satellite image classification and
clustering. 

We note that our process for converting an image into a finite dimensional
feature vector is very straightforward and does not involve any domain
knowledge about the images. In contrast to other image classification
algorithms that extract features based on sophisticated mathematical
analysis, such as, analyzing the texture, the special properties of
an image, doing edge-detection, or any of the many other methods employed
in the immense research-literature on image processing, our approach
is very basic and universal. It is based on the complexity of the
'raw' string-representation of an image. Our method extracts features
automatically just by computing distances from a set of prototypes.
It is therefore scalable and can be implemented using parallel processing
techniques, such as on system-on-chip and FPGA hardware implementation
\cite{RatsabyVadimIEEE2012,RatsabyZavielovIEEE2010,RatsabyKaspiIEEE2012}.

Our method extracts image features that are unbiased in the sense
that they do not employ any heuristics in contrast to other common
image-processing techniques\cite{canty2007image}. The features that
we extract are based on information implicit in the image and obtained
via a complexity-based UID distance which is an information-theoretic
measure. In our method, the feature vector representation of an image
is based on the distance of the image from some fixed set of representative
class-prototypes that are initially and only once picked by a human
user running the learning algorithm.

Let us now summarize the organization of the paper: in section \ref{sec:LZ-complexity-and-string}
we review the definitions of LZ-complexity and a few string distances.
In section \ref{sec:Universal-Image-Distance} we define the UID distance.
In section \ref{sec:Prototype-selection} we describe the algorithm
for selecting class prototypes. In section \ref{sec:Image-feature-representation}
we describe the algorithm that generates a feature-vector representation
of an image. In section \ref{sec:Learning-image-classification} we
discuss the classification learning method and in section we conclude
by reporting on the classification accuracy results.

\section{\label{sec:LZ-complexity-and-string}LZ-complexity and string distances}

The UID distance function \cite{RatsabyChesterIEEE2012} is based
on the LZ- complexity of a string. The definition of this complexity
follows \cite{LZ76}: let $S$,$Q$ and $R$ be strings of characters
that are defined over the alphabet ${\cal A}$. Denote by $l(S)$
the length of S, and $S(i)$ denotes the $i^{th}$ element of S. We
denote by $S(i,j)$ the substring of $S$ which consists of characters
of $S$ between position $i$ and $j$. An extension $R=SQ$ of $S$
is reproducible from $S$ (denoted as $S\rightarrow R$) if there
exists an integer $p\leq l(S)$ such that $Q(k)=R(p+k-1)$ for $k=1,\ldots,l(S)$.
For example, $aacgt\rightarrow aacgtcgtcg$ with $p=3$ and $aacgt\rightarrow aacgtac$
with $p=2$. $R$ is obtained from $S$ (the seed) by copying elements
from the $p^{th}$ location in $S$ to the end of $S$.

A string $S$ is \emph{producible} from its prefix $S(1,j)$ (denoted
$S(1,j)\Rightarrow R$), if $S(1,j)\rightarrow S(1,l(S)-1)$. For
example, $aacgt\rightarrow aacgtac$ and $aacgt\rightarrow aacgtacc$
both with pointers $p=2$. The production adds an extra 'different'
character at the end of the copying process which is not permitted
in a reproduction.

Any string $S$ can be built using a \emph{production process} where
at its $i^{th}$ step we have the production $S(1,h_{i-1})\rightarrow S(1,h_{i})$
where $h_{i}$ is the location of a character at the $i^{th}$step.
(Note that $S(1,0)\Rightarrow S(1,1)).$ 

An $m$-step production process of $S$ results in parsing of $S$
in which $H(S)=S(1,h_{1})\cdot S(h_{1}+1,h_{2})\cdots S(h_{m-1}+1,h_{m})$
is called the \emph{history} of $S$ and $H_{i}(S)=S(h_{i-1}+1,h_{i})$
is called the $i^{th}$ component of $H(S)$. For example for $S=aacgtacc$
we have $H(S)=a\cdot ac\cdot g\cdot t\cdot acc$ as the history of
$S$.

If $S(1,h_{i})$ is not reproducible from $S(1,h_{i-1})$ then $H_{i}(S)$
is called \emph{exhaustive} meaning that the copying process cannot
be continued and the component should be halted with a single character
\emph{innovation}. Moreover, every string $S$ has a unique exhaustive
history \cite{LZ76}. 

Let us denote by $c_{H}(S)$ the number of components in a history
of $S$. then the LZ complexity of $S$ is $c(S)=\min\left\{ c_{H}(S)\right\} $
where the minimum is over all histories of $S$. It can be shown that
$c(S)=c_{E}(S)$ where $c_{E}(S)$ is the number of components in the
exhaustive history of $S$.

A distance for strings based on the LZ-complexity was introduced in
\cite{Sayood2003} and is defined as follows: given two strings $X$
and $Y$, denote by $XY$ their concatenation then define 
\[
d(X,Y):=\max\left\{ c(XY)-c(X),c(YX)-c(Y)\right\} .
\]
 As in \cite{RatsabyChesterIEEE2012} we use the following normalized
distance function
\begin{equation}
d^{**}(X,Y):=\frac{c(XY)-\min\left\{ c(X),c(Y)\right\} }{\max\left\{ c(X),c(Y)\right\} }.\label{eq:d**}
\end{equation}
 We note in passing that (\ref{eq:d**}) resembles the normalized
compression distance of \cite{Cilibrasi:2005:ITIT} except that here
we do not use a compressor but rather the LZ-complexity $c$ of a
string. Note that $d^{**}$ is not a metric since it does not satisfy
the triangle inequality and a distance of $0$ implies that the two
strings are close but not necessarily identical. 

This $d^{**}$ is universal in the sense that it is not based on some
specific representation of a string (such as the alphabet of symbols),
nor on some heuristics that are common to other string distances,
e.g., edit-distances \cite{Deza-09}. Instead it only relies on the
string's LZ-complexity which is purely an information quantity independent
of the string's context or representation.

\section{\label{sec:Universal-Image-Distance}Universal Image Distance}

Based on $d^{**}$ we now define a distance between images. The idea
is to convert each of two images $I$ and $J$ into strings $X^{(I)}$
and $X^{(J)}$ of characters from a finite alphabet of symbols. Once
in string format, we use $d^{**}(X^{(I)},X^{(J)})$ as the distance
between $I$ and $J$. The details of this process are described in
Algorithm \ref{alg:UID-distance-measure} below.

\begin{algorithm}[h]
\begin{enumerate}
\item \textbf{Input}: two color images $I$, $J$ in jpeg format (RGB representation)
\item Transform the RGB matrices into gray-scale by forming a weighted sum
of the R, G, and B components according to the following formula:
$grayscaleValue:=0.2989R+0.5870G+0.1140B$, (used in Matlab\textcopyright).
Each pixel is now a single numeric value in the range of $0$ to $255$
. We refer to this set of values as the alphabet and denote it by
$\mathcal{A}$.
\item Scan each of the grayscale images from top left to bottom right and
form a string of symbols from $\mathcal{A}$. Denote the two strings
by $X^{(I)}$ and $X^{(J)}$.
\item Compute the LZ-complexities: $c\left(X^{(I)}\right)$, $c\left(X^{(J)}\right)$
and the complexity of their concatenation $c\left(X^{(I)}X^{(J)}\right)$
\item \textbf{Output}: $UID(I,J):=d^{**}\left(X^{(I)},X^{(J)}\right)$.
\end{enumerate}
\caption{\label{alg:UID-distance-measure}UID distance measure}

\end{algorithm}

In the next section we describe the process of selecting the image
prototypes.

\section{\label{sec:Prototype-selection}Prototype selection}

In this section we describe the algorithm for selecting image prototypes
from each of the feature categories . This process runs only once
before the stage of converting the images into finite dimensional
vectors, that is, it does not run once per image but once for all
images. For an image $I$ we denote by $P\subset I$ a sub-image $P$
of $I$ where $P$ can be any rectangular-image obtained by placing
a window over the image $I$ where the window is totally enclosed
by $I$.

\begin{algorithm}[H]
\begin{enumerate}
\item \textbf{Input}: $M$ image feature categories, and a corpus $\mathcal{C}_{N}$
of $N$ unlabeled colored images $\left\{ I_{j}\right\} _{j=1}^{N}$
.
\item \textbf{for ($i:=1$ to $M$) do}

\begin{enumerate}
\item Based on \emph{any} of the images $I_{j}$ in $\mathcal{C}_{N}$,
let the user \textbf{select} $L_{i}$ prototype images $\left\{ P_{k}^{(i)}\right\} _{k=1}^{L_{i}}$
and set them as feature category $i$. Each prototype is contained
by some image, $P_{k}^{(i)}\subset I_{j}$, and the size of $P_{k}^{(i)}$
can vary, in particular it can be much smaller than the size of the
images $I_{j}$, $1\leq j\leq N$.
\item \textbf{end for;}
\end{enumerate}
\item \textbf{Enumerate} all the prototypes into a single \emph{unlabeled}
set $\left\{ P_{k}\right\} _{k=1}^{L}$, where $L=\sum_{i=1}^{M}L_{i}$
and calculate the distance matrix $H=\left[UID\left(X^{(P_{k})},X^{(P_{l})}\right)\right]_{k=1,l=1}^{L}$
where the $(k,l)$ component of $H$ is the UID distance between the
unlabeled prototypes $P_{k}$ and $P_{l}$.
\item \textbf{Run} hierarchical clustering on $H$ and obtain the associated
dendrogram.
\item \textbf{If} there are $M$ clusters with the $i^{th}$ cluster consisting
of the prototypes $\left\{ P_{k}^{(i)}\right\} _{k=1}^{L_{i}}$ \textbf{then}
terminate and \textbf{go to} step $7$. 
\item \textbf{Else go to }step 2\textbf{.}
\item \textbf{Output}: the set of labaled prototypes $\mathcal{P}_{L}:=\left\{ \left\{ P_{k}^{(i)}\right\} _{k=1}^{L_{i}}\right\} _{i=1}^{M}$
where $L$ is the number of prototypes.
\end{enumerate}
\caption{\label{alg:prototype-sel}Prototypes selection}
\end{algorithm}

From the theory of learning pattern recognition, it is known that
the dimensionality $M$ of a feature-vector is usually taken to be
small compared to the data size $N$. A large $L$ will obtain better
feature representation accuracy of the image, but it will increase
the time for running Algorithm \ref{alg:Case-generation} (described
below).

Algorithm \ref{alg:prototype-sel} convergence is based on the user's
ability to select good prototype images. We note that from our experiments
this is easily achieved primarily because the UID permits to select
prototypes $P_{k}^{(i)}$ which are considerably smaller in size and
hence simpler than the full images $I_{j}^{(i)}$. For instance, in
our experiments we used $45\times1$7 pixels prototype size for all
feature  categories. This fact makes it easy for a user to quickly
choose typical representative prototypes from every feature-cateory.
This way it is easy to find informative prototypes, that is, prototypes
that are distant when they are from different feature-categories and
close when they are from the same feature category. Thus Algorithm
\ref{alg:prototype-sel} typically converges rapidly.

\begin{figure}[h]
\begin{centering}
\subfloat[\label{fig:Prototypes-of-categories}Labeled \emph{p}rototypes of
feature-categories\emph{ urban}, \emph{sea ,} \emph{roads}, and \emph{arid}
(each feature has three prototypes, starting from top left and moving
right in sequence)]{\begin{centering}
\includegraphics{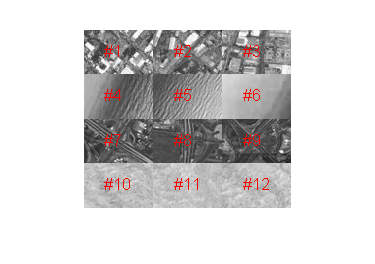}
\par\end{centering}

}
\par\end{centering}

\begin{centering}
\subfloat[\label{fig:Dendogram-produced-in}Dendrogram produced in step 4 of
Algorithm \ref{alg:prototype-sel}.]{\begin{centering}
\includegraphics[scale=0.5]{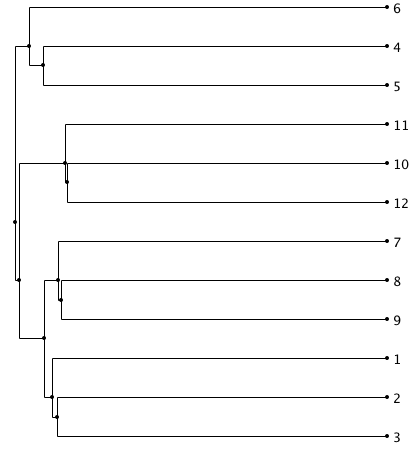}
\par\end{centering}

}
\par\end{centering}

\caption{Prototypes and dendrogram of Algorithm \ref{sec:Prototype-selection}}
\end{figure}

As an example, Figure \ref{fig:Prototypes-of-categories} displays
$12$ prototypes selected by a user from a corpus of satellite images.
The user labeled prototypes $1,\ldots,3$ as representative of the
feature category \emph{urban,} prototypes $4,\ldots,6$ as representatives
of class \emph{sea, }prototypes\emph{ $7,\ldots,9$ }as representative
of feature \emph{roads} and prototypes $10,\ldots,12$ as representative
of feature \emph{arid. }The user easily found these representative
prototypes as it is easy to fit in a single picture of size $45\times17$
pixels a typical image. The dendrogram produced in step 4 of Algorithm
\ref{alg:prototype-sel} for these set of $12$ prototypes is displayed
in Figure \ref{fig:Dendogram-produced-in}. It is seen that the following
four clusters were found $\left\{ 10,12,11\right\} ,\left\{ 1,2,3\right\} ,\left\{ 7,8,9\right\} ,\left\{ 4,6,5\right\} $
which indicates that the prototypes selected in Algorithm \ref{alg:prototype-sel}
are good.

\section{\label{sec:Image-feature-representation}Image feature-representation}

In the previous section we described Algorithm \ref{alg:prototype-sel}
by which the prototypes are manually selected. This algorithm is now
used to create a feature-vector representation of an image. It is
described as Algorithm \ref{alg:Case-generation} below (in \cite{RatsabyChesterIEEE2012}
we used a similar algorithm UIC to soft-classify an image whilst here
we use it to only produce a feature vector representation of an image
which later serves as a single labeled case for training any supervised
learning algorithm or a single unlabeled case for training an unsupervised
algorithm).

\begin{algorithm}[H]
\begin{enumerate}
\item \textbf{Input}: an image $I$ to be represented on the following feature
categories $1\leq i\leq M$, and given a set $\mathcal{P}_{L}:=\left\{ \left\{ P_{k}^{(i)}\right\} _{k=1}^{L_{i}}\right\} _{i=1}^{M}$
of labeled prototype images (obtained from Algorithm \ref{alg:prototype-sel}).
\item \textbf{Initialize} the count variables $c_{i}:=0$, $1\leq i\leq M$ 
\item Let $W$ be a rectangle of size equal to the maximum prototype size.
\item Scan a window $W$ across $I$ from top-left to bottom-right in a
non-overlapping way, and let the sequence of obtained sub-images of
$I$ be denoted as $\left\{ I_{j}\right\} _{j=1}^{m}$. 
\item \textbf{for} ($j:=1$ to $m$) \textbf{do}

\begin{enumerate}
\item \textbf{for} ($i:=1$ to $M$) \textbf{do}

\begin{enumerate}
\item $temp:=0$
\item \textbf{for} ($k:=1$ to $L_{i}$) \textbf{do}

\begin{enumerate}
\item $temp:=temp+\left(UID(I_{j},P_{k}^{(i)})\right)^{2}$
\item \textbf{end for;}
\end{enumerate}
\item $r_{i}:=\sqrt{temp}$
\item \textbf{end for;}
\end{enumerate}
\item Let $i^{*}(j):=\text{argmin}_{1\leq i\leq M}r_{i}$, this is the decided
feature category for sub-image $I_{j}$.
\item \textbf{Increment }the count, $c_{i^{*}(j)}:=c_{i^{*}(j)}+1$ 
\item \textbf{end for;}
\end{enumerate}
\item \textbf{Normalize} the counts, $v_{i}:=\frac{c_{i}}{\sum_{l=1}^{M}c_{l}}$,
$1\leq i\leq M$ 
\item \textbf{Output}: the normalized vector $v(I)=\left[v_{1},\ldots v_{M}\right]$
as the feature-vector representation for image $I$
\end{enumerate}
\caption{\label{alg:Case-generation}Feature-vector generation}
\end{algorithm}

\section{\label{sec:Learning-image-classification}Supervised and unsupervised
learning on images}

Given a corpus ${\cal C}$ of images and a set ${\cal P}_{L}$ of
labeled prototypes we use Algorithm \ref{alg:Case-generation} to
generate the feature-vectors $v(I)$ corresponding to each image $I$
in ${\cal C}$. At this point we have a database ${\cal D}$ of size
equal to $|{\cal C}|$ which consists of feature vectors of all the
images in ${\cal C}$. This database can be used for \emph{unsupervised}
learning, for instance, discover interesting clusters of images. It
can also be used for \emph{supervised} learning provided that each
of the cases can be labeled according to a value of some target class
variable which in general may be different from the feature categories.
Let us denote by $T$ the class target variable and the database ${\cal D}_{T}$
which consists of the feature vectors of ${\cal D}$ with the corresponding
target class values. The following 

\begin{algorithm}[h]
\begin{enumerate}
\item \textbf{Input}: (1) a target class variable $T$ taking values in
a finite set $\mathcal{T}$ of class categories, (2) a database ${\cal D}_{T}$
which is based on the $M$-dimensional feature-vectors database ${\cal D}$
labeled with values in $\mathcal{T}$ (3) any supervised learning
algorithm $\mathcal{A}$
\item Partition ${\cal D}_{T}$ using $n$-fold cross validation into Training
and Testing sets of cases
\item Train and test algorithm $\mathcal{A}$ and produce a classifier $C$
which maps the feature space $[0,1]^{M}$ into $\mathcal{T}$
\item Define Image classifier as follows: given any image $I$ the classification
is $F(I):=C(v(I))$, where $v(I)$ is the $M$-dimensional feature
vector of $I$
\item \textbf{Output}: classifier $F$
\end{enumerate}
\caption{\label{alg:Image-classification-learning}Image classification learning}
\end{algorithm}

\section{\label{sec:Experimental-setup-and}Experimental setup and results}

\begin{figure}
\subfloat[]{\includegraphics[scale=0.15]{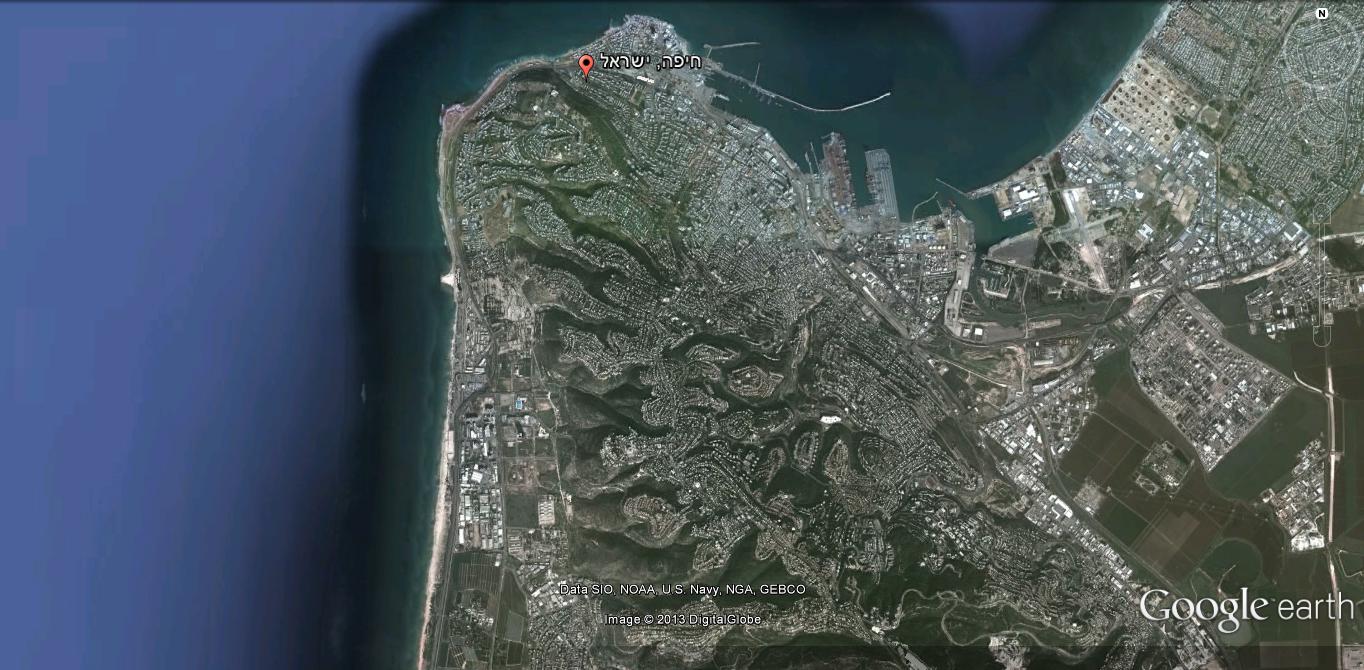}

}\subfloat[]{\includegraphics[scale=0.15]{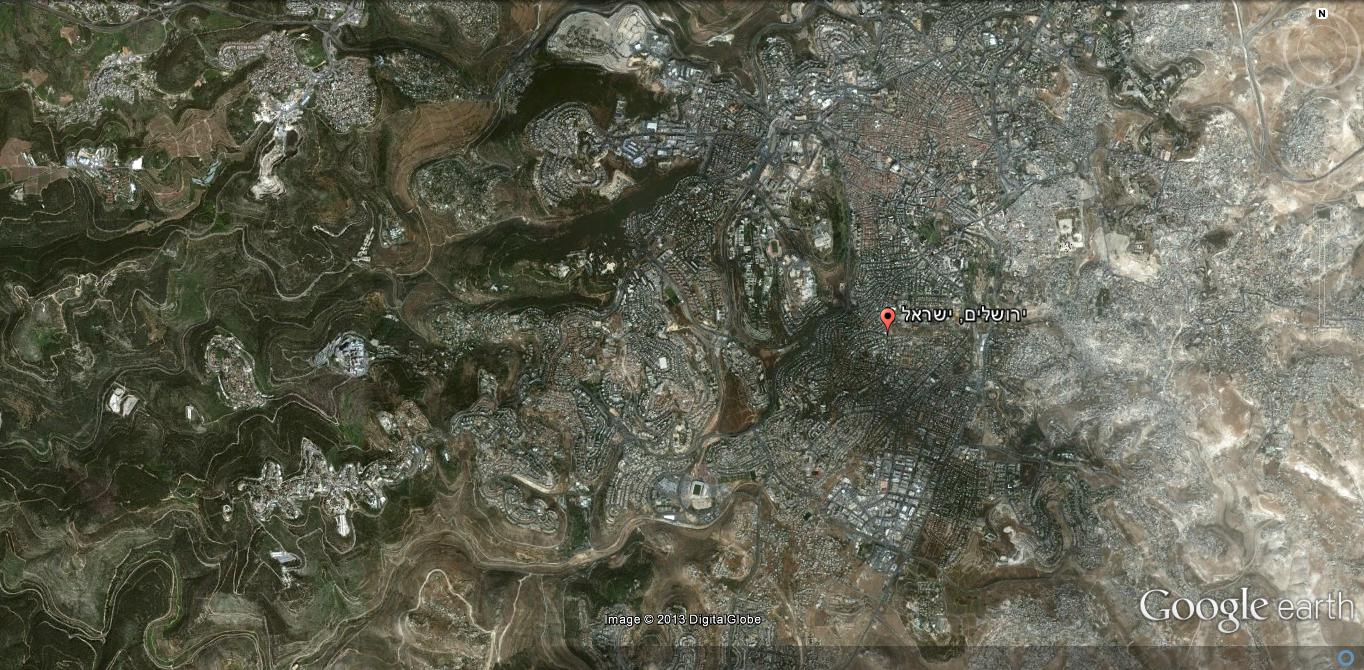}

}

\subfloat[]{\includegraphics[scale=0.15]{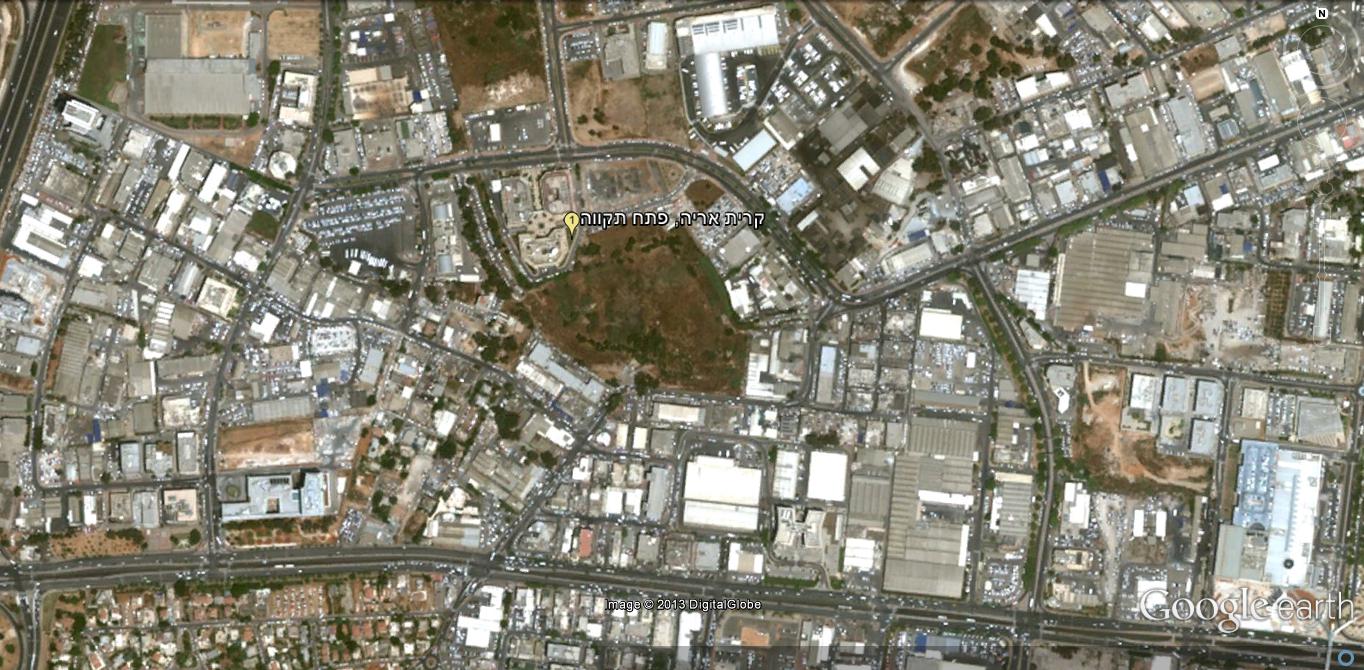}

}\subfloat[]{\includegraphics[scale=0.15]{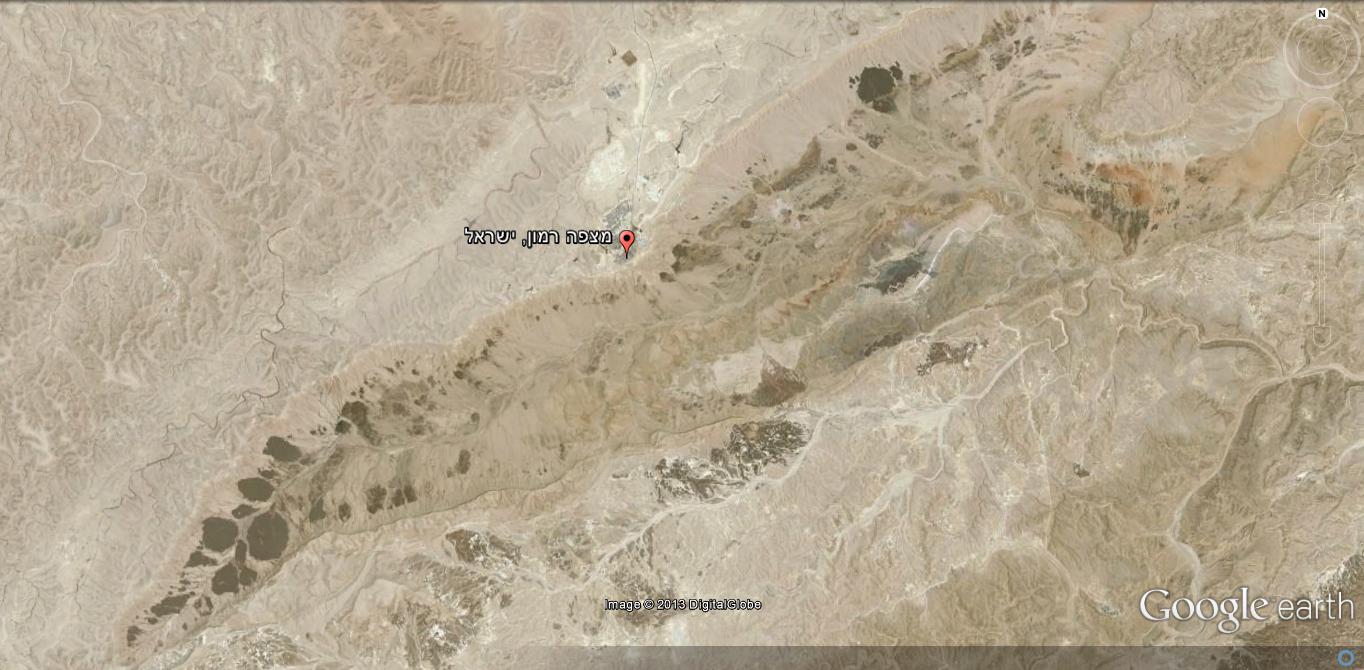}

}

\subfloat[]{\includegraphics[scale=0.15]{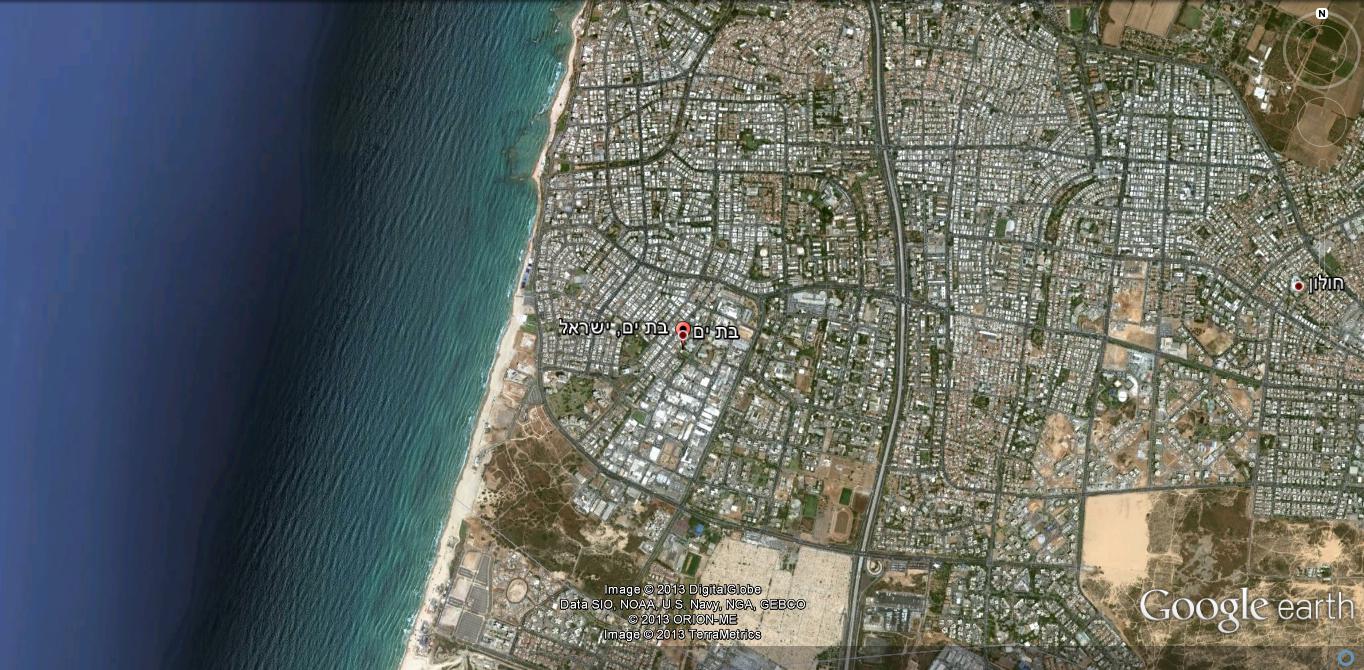}

}\subfloat[]{\includegraphics[scale=0.15]{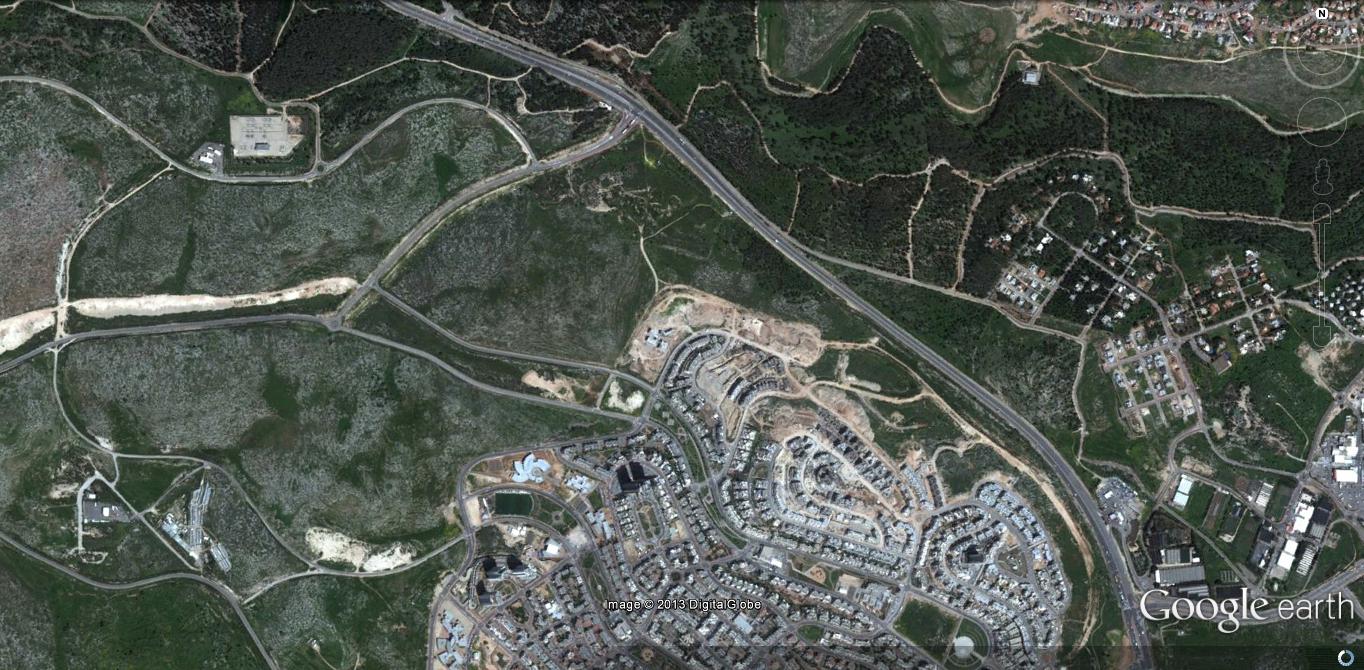}

}

\subfloat[]{\includegraphics[scale=0.15]{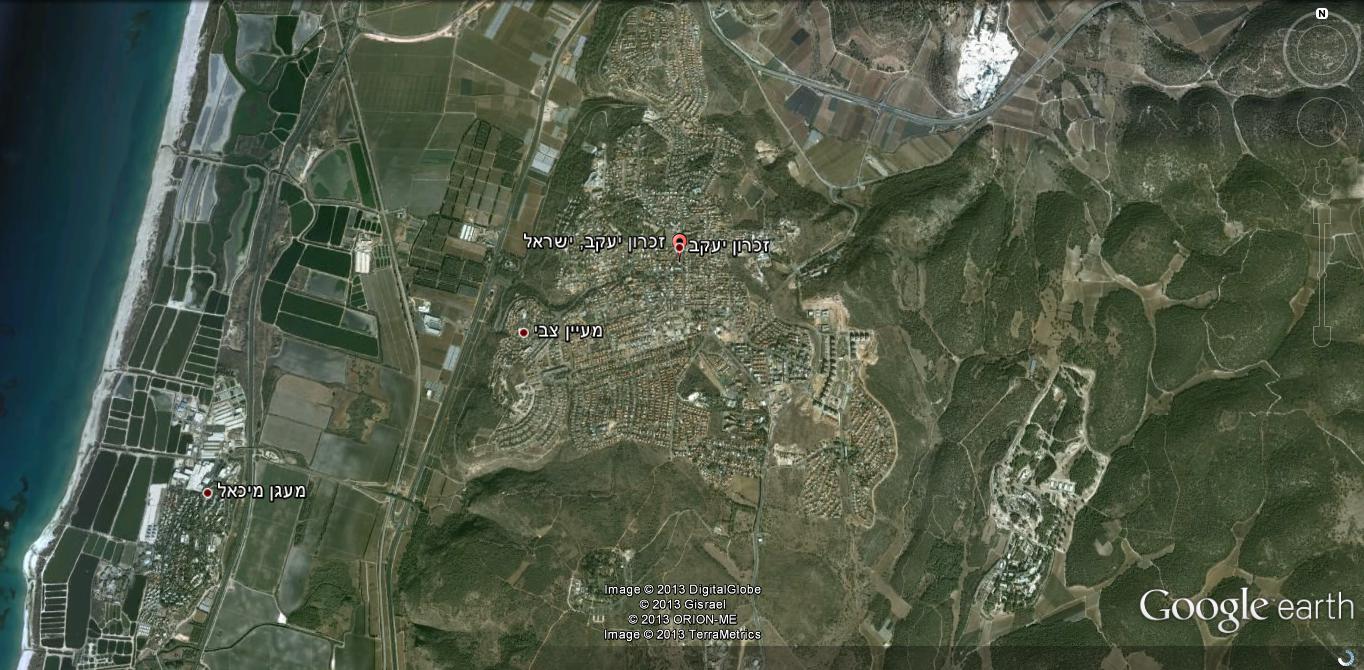}

}\subfloat[]{\includegraphics[scale=0.15]{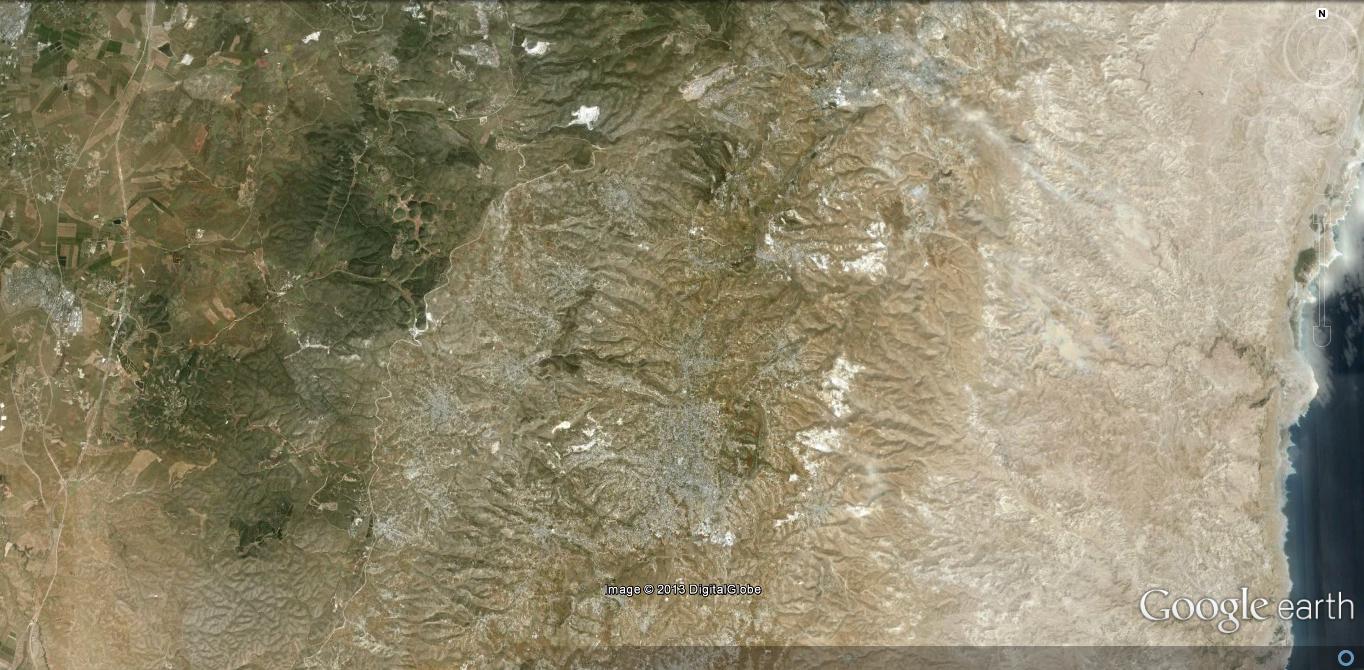}

}

\caption{\label{fig:Examples-of-images}Examples of images in the corpus}

\end{figure}

We created a corpus $\mathcal{C}$ of $60$ images of size $670\times1364$
pixels from GoogleEarth\textcopyright of various types of areas (Figure \ref{fig:Examples-of-images}
displays a few scaled-down examples of such images). From these images
we let a user define four feature-categories: \emph{sea}, \emph{urban},
\emph{arid}, \emph{roads} and choose three relatively-small image-prototype
of size $45\times17$ pixels from each feature-category, that is,
we ran Algorithm \ref{alg:prototype-sel} with $M=4$ and $L_{i}=3$
for all $1\leq i\leq M$. We then ran Algorithm \ref{alg:Case-generation}
to generate the feature-vectors for each image in the corpus and obtained
a database ${\cal D}$.

We then let the user label the images by a target variable \emph{Humidity}
with possible values $0$ or $1$. An image is labeled $0$ if the
area is of low humidity and labeled $1$ if it is of higher humidity.
We note that an image of a low humidity region may be in an arid (dry)
area or also in the higher-elevation areas which are not necessarily
arid. Since elevation information is not available in the feature-categories
that the user has chosen then the classification problem is hard since
the learning algorithm needs to discover the dependency between humid
regions and areas characterized only by the above four feature categories.

With this labeling information at hand we produced the labeled database
${\cal D}_{Humidity}$. We used Algorithm \ref{alg:Image-classification-learning}
to learn an image classifier with target \emph{Humidity.} As the learning
algorithm ${\cal A}$ we used the following standard supervised algorithms:
$J48$, $CART$, which learn decision trees, \emph{NaiveBayes} and
\emph{Multi-Layer Perceptrons} (backpropagation) all of which are
available in the WEKA\textcopyright toolkit.

We performed $10$-fold cross validation and compared their accuracies
to a baseline classifier (denoted as ZeroR) which has a single decision
that corresponds to the class value with the highest \emph{prior}
empirical probability. As seen in Table \ref{results1} (generated
by WEKA\textcopyright) $J48$, \emph{CART}, NaiveBayes and Backpropagation performed
with an accuracy of $86.5\%$, $81.5\%$, $89.25\%$, and $87.25\%$,
respectively, compared to $50\%$ achieved by the baseline $ZeroR$
classifier. The comparison concludes that all three learning algorithms
are significantly better than the baseline classifier, based on a
T-test with a significance level of $0.05$.

\begin{table}[thb] \footnotesize {\centering \begin{tabular}{lrr@{\hspace{0.1cm}}cr@{\hspace{0.1cm}}cr@{\hspace{0.1cm}}cr@{\hspace{0.1cm}}c} \\ \hline Dataset  ${\cal D}_{Humidity}$& (1)& (2) & & (3) & & (4) & & (5) & \\
\hline \hline Classify Image into Humidity:\hspace{0.1cm}  & 50.00 & 86.50 & $\circ$ & 81.50 & $\circ$ & 89.25 & $\circ$ & 87.25 & $\circ$\\ \hline \multicolumn{10}{c}{$\circ$, $\bullet$ statistically significant improvement or degradation}\\ \end{tabular} \footnotesize \par} 
\scriptsize {\centering \begin{tabular}{cl}\\ (1) & rules.ZeroR '' 48055541465867954 \\ (2) & trees.J48 '-C 0.25 -M 2' -217733168393644444 \\ (3) & trees.SimpleCart '-S 1 -M 2.0 -N 5 -C 1.0' 4154189200352566053 \\ (4) & bayes.NaiveBayes '' 5995231201785697655 \\ (5) & functions.MultilayerPerceptron '-L 0.3 -M 0.2 -N 500 -V 0 -S 0 -E 20 -H a' -5990607817048210779 \\ \end{tabular} }
\caption{\label{results1}Percent correct results for classifying {\em Humidity}}
\end{table}

Next, we performed clustering on the unlabeled database ${\cal D}$.
Using the k-means algorithm, we obtained 3 significant clusters, shown
in Table \ref{tab:EM-Clusters}.
\begin{table}
\begin{centering}
\begin{tabular}{|>{\centering}m{2.5cm}|>{\centering}m{2.5cm}|>{\centering}m{2cm}|>{\centering}m{2cm}|>{\centering}m{2cm}|}
\hline 
Feature & Full data & Cluster\#1 & Cluster\#2 & Cluster\#3\tabularnewline
\hline 
urban & 0.3682 & 0.6219 & 0.1507 & 0.2407\tabularnewline
\hline 
sea & 0.049 & 0.0085 & 0 & 0.1012\tabularnewline
\hline 
road & 0.4074 & 0.2873 & 0.0164 & 0.655\tabularnewline
\hline 
arid & 0.1754 & 0.0824 & 0.8329 & 0.003\tabularnewline
\hline 
\end{tabular}
\par\end{centering}

\caption{\label{tab:EM-Clusters}k-means clusters found on unsupervised database
${\cal D}$}

\end{table}
 The first cluster captures images of highly urban areas that are
next to concentration of roads, highways and interchanges while the
second cluster contains less populated (urban) areas in arid locations
(absolutely no sea feature seen) with very low concentration of roads.
The third cluster captures the coastal areas and here we can see that
there can be a mixture of urban (but less populated than images of
the first cluster) with roads and extremely low percentage of arid
land.

The fact that such interesting knowledge can be extracted from raw
images using our feature-extraction method is very significant since
as mentioned above our method is fully automatic and requires no image
analysis or any sophisticated preprocessing stages that are common
in image pattern analysis.

\section{Conclusion}

We introduced a method for automatically defining and measuring features
of colored images.The method is based on a universal image distance
that is measured by computing the complexity of the string-representation
of the two images and their concatenation. An image is represented
by a feature-vector which consists of the distances from the image
to a fixed set of small image prototypes, defined once by a user.
There is no need for any sophisticated mathematical-based image analysis
or pre-processing since the universal image distance regards the image
as a string of symbols which contains all the relevant information
of the image. The simplicity of our method makes it very attractive
for fast and scalable implementation, for instance on a specific-purpose
hardware acceleration chip. We applied our method to supervised and
unsupervised machine learning on satellite images. The results show
that standard machine learning algorithms perform well based on our
feature-vector representation of the images.

\bibliographystyle{plain}

\end{document}